\documentclass[runningheads]{llncs}
\usepackage[T1]{fontenc}
\usepackage[frozencache=true,cachedir=minted-cache]{minted} 
\usepackage{amsmath}
\usepackage{bbding}
\usepackage{amsfonts}
\usepackage{algorithm}
\usepackage{algorithmic}
\usepackage{graphicx}
\usepackage{booktabs} 
\usepackage{makecell}
\usepackage[colorlinks, linkcolor=blue, urlcolor=blue,
anchorcolor=blue, citecolor=blue]{hyperref}
\usepackage{multirow}
\usepackage{color,xcolor}
\usepackage{array}
\usepackage{url}
\usepackage{pifont}

\definecolor{red}{RGB}{255,0,0}

\begin{document}

\title{Treasure in Distribution: A Domain Randomization based Multi-Source Domain Generalization for 2D Medical Image Segmentation
}

\titlerunning{Treasure in Distribution}
%
\author{
Ziyang Chen \inst{1}\thanks{\small Z. Chen and Y. Pan contributed equally. Corresponding author: Y. Xia.}
\and
Yongsheng Pan \inst{2}$^{\star}$
\and
Yiwen Ye \inst{1}
\and
Hengfei Cui \inst{1}
\and
Yong Xia\inst{1} 
}

\authorrunning{Z. Chen et al.}

%

\institute{National Engineering Laboratory for Integrated Aero-Space-Ground-Ocean Big Data Application Technology, School of Computer Science and Engineering, Northwestern Polytechnical University, Xi’an 710072, China \\ \and
School of Biomedical and Engineering, ShanghaiTech University, Shanghai 201210, China\\
\email{zychen@mail.nwpu.edu.cn, yspan@mail.nwpu.edu.cn, ywye@mail.nwpu.edu.cn, hfcui@nwpu.edu.cn, yxia@nwpu.edu.cn} \\
}

\maketitle              
\begin{abstract}
Although recent years have witnessed the great success of convolutional neural networks (CNNs) in medical image segmentation, the domain shift issue caused by the highly variable image quality of medical images hinders the deployment of CNNs in real-world clinical applications. 
Domain generalization (DG) methods aim to address this issue by training a robust model on the source domain, which has a strong generalization ability.
Previously, many DG methods based on feature-space domain randomization have been proposed, which, however, suffer from the limited and unordered search space of feature styles.
In this paper, we propose a multi-source DG method called \textbf{Tr}easure \textbf{i}n \textbf{D}istribution (TriD), which constructs an unprecedented search space to obtain the model with strong robustness by randomly sampling from a uniform distribution. To learn the domain-invariant representations explicitly, we further devise a style-mixing strategy in our TriD, which mixes the feature styles by randomly mixing the augmented and original statistics along the channel wise and can be extended to other DG methods.
Extensive experiments on two medical segmentation tasks with different modalities demonstrate that our TriD achieves superior generalization performance on unseen target-domain data.
Code is available at \href{https://github.com/Chen-Ziyang/TriD}{https://github.com/Chen-Ziyang/TriD}.

\keywords{Domain generalization  \and Domain randomization \and Medical image segmentation \and Deep learning.}
\end{abstract}
%
%


\section{Introduction}
Medical image segmentation is an essential task in computer-aided diagnosis. On this task, convolutional neural networks (CNNs) have demonstrated their effectiveness in an extensive literature~\cite{SegSurvey1,SegSurvey2}. 
However, these CNN models obtained on training (\emph{i.e.}, source domain) data can hardly generalize well on the unseen test (\emph{i.e.}, target domain) data.
The poor generalization ability, which hinders CNNs to be used in real-world clinical applications, can be attributed to the fact that the quality of medical images varies greatly across healthcare centers with different scanners and imaging protocols, resulting in large distribution discrepancy (\emph{a.k.a.}, domain shift). 
To improve the generalization ability, domain generalization (DG) methods have been proposed~\cite{DGSurvey1,DGSurvey2}. These methods can be trained on the data from one or multiple source domains. Considering the diversity of training data, we focus on multi-source DG in this study.

Most studies on DG attempt to alleviate the distribution discrepancy by standardizing the features~\cite{IBN,SW,ISW,SAN-SAW} and/or adding extra structures~\cite{DCAC,BNE} to the network. 
However, the former suffers from over-standardization and may hinder the network to preserve semantic contents, while the latter may introduce excess misjudgment risk when estimating the distance between source- and target-domain data.


\begin{figure}[!t]
	\centering
	\includegraphics[width=0.8\textwidth]{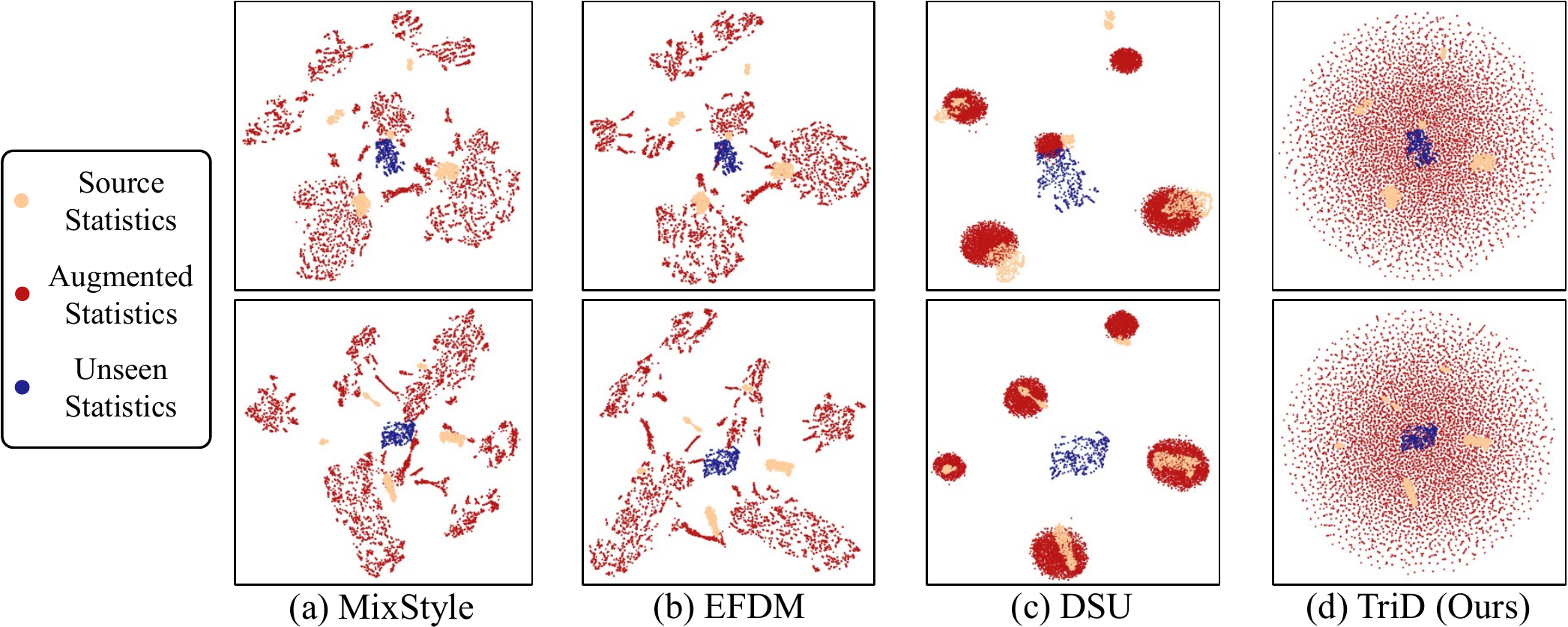}
	\caption{Visualization of the statistics (1st row: standard deviation, 2nd row: mean) computed from the features of the first residual block of ResNet-34 trained on prostate dataset, which has six different domains. We take five as source domains and the left one as unseen target domain and visualize the augmented statistics produced by MixStyle, EFDM, DSU and our SR using 2D t-SNE. (a) MixStyle: Using a linear combination of different statistics. (b) EFDM: Using exact histogram matching and the statistics-fusion operation of MixStyle. (c) DSU: Sampling from a normal distribution constructed based on the original statistics. (d) TriD (Ours): Sampling from a uniform distribution. }
	\label{fig-tsne}
\end{figure}

A recent mainstream in DG research is to simulate the distributions of unseen target-domain data via domain randomization, \emph{i.e.}, perturbing the styles of source-domain data. 
The perturbation function can be defined in the input space~\cite{RandConv,Geo,GTR-LTR}. Thus, it is easy to evaluate the quality of perturbed images, but the definition generally requires domain knowledge and expertise~\cite{MaxStyle}.
By contrast, the perturbation can be performed in the feature space~\cite{MixStyle,DSU,EFDM}. However, this may cause difficulties in monitoring the perturbation degree of semantic contents in the feature space due to the lack of visualization.
Recently, two critical attributes of the feature space are revealed by IBN-Net~\cite{IBN} and AdaIN~\cite{AdaIN}, respectively. First, most style-texture information resides in the low-level features extracted by shallow layers. Second, the content-preserving style transformation can be performed by changing the statistics (\emph{e.g.}, mean and standard deviation) of the low-level features. 
Inspired by them, MixStyle~\cite{MixStyle} perturbs the feature styles using augmented statistics, which are generated by randomly mixing the statistics of the low-level features from two samples.
Subsequently, more research efforts have been devoted to designing the search space that covers a larger area in the feature-style space~\cite{DSU,EFDM}.
Despite their improved performance, using the statistics of source-domain data for feature perturbation may limit the search space and can hardly explore in the feature-style space evenly (see Fig.~\ref{fig-tsne}(a), (b) and (c)).
The points indicating the augmented statistics are scattered and do not completely cover the points from unseen target domain.
Moreover, since all feature channels are perturbed, these methods lack a reference to the original feature, which prevents them from learning the domain-invariant representations explicitly.

To address these issues, in this paper, we propose a simple but effective multi-source DG method called \textbf{Tr}easure \textbf{i}n \textbf{D}istribution (TriD), which consists of two major steps: statistics randomization (SR) and style mixing (SM). 
SR aims to tap the potential of distribution by randomly sampling the augmented statistics from a uniform distribution to perturb the original intermediate features, which can expand the search space to cover more cases evenly (see Fig.~\ref{fig-tsne}(d)).
It can be observed that the red points are distributed evenly and cover not only the unseen target domain, but also the source domains.
This leads to the issue that the perturbed features may have unreal styles. We hypothetically extend the unreal styles to the feature space with the inspiration from~\cite{GTR-LTR}, which demonstrated the effectiveness of unreal styles in the input space. 
SM is devised to mix the feature styles by randomly mixing the augmented and original statistics in the channel dimension, thus making it feasible to learn the domain-invariant representations explicitly.
We have evaluated our proposed TriD on two medical segmentation tasks: (1) the prostate segmentation using magnetic resonance imaging (MRI) from six domains and (2) joint segmentation of optic disc (OD) and optic cup (OC) in fundus images from five domains.
Extensive experiments demonstrate that our TriD achieves a superior generalization ability to the state-of-the-art DG methods on unseen target-domain data.

Our contributions are three-fold:
(1) The proposed multi-source DG method called TriD can boost the robustness of model and alleviate the performance drop on the unseen target-domain data.
(2) We focus on expanding the search space of feature styles and therefore devise the statistics-randomization strategy, which allows exploring in the feature-style space evenly.
(3) Different from perturbing all feature channels, we introduce the original statistics to the augmented statistics to learn the domain-invariant representations explicitly.

\section{Method}
\subsection{Preliminaries}
Let $f \in R^{B\times C\times H\times W}$ be the intermediate features in a mini-batch, where $B$, $C$, $H$, and $W$ respectively denote the mini-batch size, channel, height, and width. MixStyle~\cite{MixStyle} perturbs the features by randomly mixing different feature statistics, formulated as follows:

\begin{equation}
    \begin{aligned}
        MixStyle(f_i) = \gamma_{m} \frac{f_i-\mu(f_i)}{\sigma(f_i)}+\beta_{m},
    \end{aligned}
    \label{eq-mixstyle}
\end{equation}

\begin{equation}
    \begin{aligned}
        \gamma_{m} = \lambda_{m}\sigma(f_i) + (1-\lambda_{m})\sigma(f_j),\; \beta_{m} = \lambda_{m}\mu(f_i) + (1-\lambda_{m})\mu(f_j),
    \end{aligned}
\end{equation}

where $\lambda_{m}\in R^B$ is a weight coefficient sampled from a Beta distribution~\cite{MixStyle}, $f_i,f_{j(j\ne i)}\in R^{C\times H \times W}$ indicate the features from two different images in a mini-batch, and $\mu(*),\sigma(*)\in R^{B\times C}$ are the mean and standard deviation computed across the spatial dimension within each channel of each image. 

\begin{figure}[!tb]
	\centering
	\includegraphics[width=0.8\textwidth]{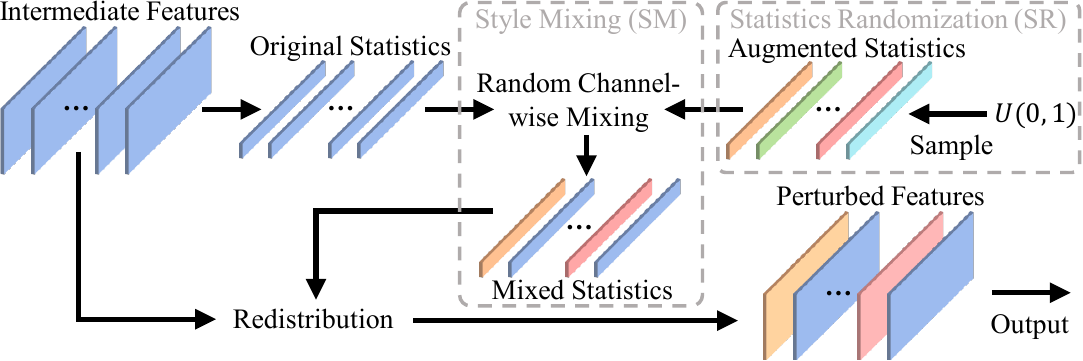}
	\caption{Overview of our TriD for feature perturbation. 
 }
	\label{fig-overview}
\end{figure}

\subsection{Treasure in Distribution (TriD)}
The TriD is designed to perturb the intermediate feature styles by randomly changing the feature statistics (\emph{i.e.}, mean and standard deviation), as shown in Fig.~\ref{fig-overview}.
The feature statistics is substituted by the mixed statistics, which is generated by mixing the augmented and original statistics in channel dimension.
It can be implemented as a plug-and-play module inserted into any CNN-based architecture. In this study, we use ResNet-34~\cite{ResNet} as the backbone to construct the segmentation network in a U-shape architecture~\cite{DoCR}. The TriD is inserted behind the first and second residual blocks during training, and will be removed in the inference phase. We now delve into the details of our TriD.

\noindent{\textbf{Statistics Randomization (SR).}}
Inspired by effectiveness of unreal styles in the input space~\cite{GTR-LTR}, we hypothesis that unreal styles can also be extended to the feature space. To cover more cases evenly, we randomly sample the augmented statistics $\sigma_r,\mu_r \in R^{B \times C}$ from a uniform distribution which contains most feature statistics: $\sigma_r\sim U(0,1)$, $\mu_r\sim U(0,1)$.

\noindent{\textbf{Style Mixing (SM).}}
To learn the domain-invariant representations explicitly, SM strategy is designed to randomly mix the augmented and original statistics along the channel wise. We first sample $P\in R^{B\times C}$ from the Beta distribution: $P \sim Beta(\alpha, \alpha)$, 
and use $P$ as the probability to generate the Bernoulli distribution from which to sample $\lambda \in R^{B\times C}$: $\lambda \sim Bern(P)$, where $\alpha$ is set to $0.1$ empirically~\cite{MixStyle}.
Then the mixed statistics is calculated as:
\begin{equation}
    \begin{aligned}
        \gamma_{mix} = \lambda \sigma(r) + (1 - \lambda) \sigma(f), \;
        \beta_{mix} = \lambda \mu(r) + (1 - \lambda) \mu(f),
    \end{aligned}
\end{equation}
where $f$ denotes the intermediate features.
Finally, the mixed feature statistics is applied to perturb the normalized $f$ similar to Eq.~(\ref{eq-mixstyle}),
\begin{equation}
    \begin{aligned}
        TriD(f) = \gamma_{mix} \frac{f-\mu(f)}{\sigma(f)}+\beta_{mix}.
    \end{aligned}
\end{equation}
Different from MixStyle, we replace the batch-wise fusion with channel-wise mixing, which avoids the sampling preference and introduces original-feature reference, so as to learn the domain-invariant representations explicitly.

\subsection{Training and Inference}
Let $\mathcal{D}_s = \{(x_{di}, y_{di})^{N_d}_{i=1} \}^K_{d=1}$ be a set including K source domains, where $x_{di}$ is the $i$-th image in the $d$-th source domain, and $y_{di}$ is the corresponding segmentation mask of $x_{di}$. Our goal is to train a segmentation model that can generalize well
to an unseen target domain $\mathcal{D}_t = (x_{i})^{N_t}_{i=1}$.

\noindent{\textbf{Training.}} During training, we empirically set a probability of $0.5$ to activate TriD in the forward pass~\cite{MixStyle}. The segmentation network is trained on the source domains $\mathcal{D}_s$ by using the combination of Dice loss ($\mathcal{L}_{Dice}$) and cross-entropy loss ($\mathcal{L}_{ce}$) as the objective: 
$\mathcal{L}_{seg} = \mathcal{L}_{Dice}+\mathcal{L}_{ce}$.

\noindent{\textbf{Inference.}} During inference, all the TriD modules are removed, and the segmentation network is tested on the unseen target domain $\mathcal{D}_t$.


\section{Experiments and Results}
\subsection{Datasets and Evaluation Metrics}
Two datasets are used for this study, whose details are summarized in Table~\ref{tab:dataset}. 

The first dataset contains 116 MRI cases from six domains for prostate segmentation~\cite{Prostate}. We preprocess these MRI cases same as a previous study~\cite{DCAC} and only preserve the slices with the prostate region for consistent and objective segmentation evaluation. These slices are resized to 384$\times$384 with same voxel spacing. On this dataset, we employ the Dice Similarity Coefficient (DSC) and Average Surface Distance (ASD) to evaluate the prostate segmentation. Note that we regard prostate segmentation as a 2D segmentation task, but calculate metrics on 3D volumes.

The second dataset is a collection of two large and three small public datasets used for joint segmentation of optic disc (OD) and optic cup (OC)~\cite{RIGA,REFUGE,ORIGA,Drishti-GS}, which can evaluate our TriD under different data-amount scenarios. Each of these five datasets has a training/test split, and in total, we have 1,102 cases for training and 339 cases for test. Each image is center-cropped and resized to 512$\times$512~\cite{DoCR}. We employ DSC to evaluate the joint segmentation of OD and OC.



\subsection{Implementation Details}
We set the mini-batch size to 8 and adopt the SGD optimizer with a momentum of 0.99 for both tasks. The initial learning rate $l_0$ is set to 0.01 (prostate) and 0.001 (OD/OC) respectively and decays according to the polynomial rule $l_t = l_0 \times (1-t/T)^{0.9}$, where $l_t$ is the learning rate of the $t$-th epoch and $T$ is the number of total epochs that is set to 200 for prostate segmentation and 100 for joint segmentation of OD and OC. 
For both tasks, 
the leave-one-domain-out strategy was used to evaluate the performance of each DG method, \emph{i.e.}, training on $K$-1 source domains and evaluating on the left domain. We consistently apply the above implementation settings to our TriD and other competing methods.

\begin{table}[!t]
  \centering
  \caption{Details of two datasets used for this study. Number of cases with `/' denotes the data split (training/test cases).}
  \resizebox{1\textwidth}{!}{
    \begin{tabular}{ccccccc}
    \toprule
    Task & & Modality & & Number of Domains & & Cases in Each Domain\\
    \midrule
    Prostate Segmentation & & MRI & & 6 & & 30; 30; 19; 13; 12; 12\\
    OD/OC Segmentation & & Color Fundus Image & & 5 & & 156/39; 76/19; 320/80; 500/150; 50/51 \\
    \bottomrule
    \end{tabular}
    }
  \label{tab:dataset}
\end{table}

\subsection{Results}
\begin{table}[!tb]
    \caption{Performance (OD, OC) of Intra-Domain, DeepAll, our TriD and seven DG methods in joint segmentation of OD and OC. The best results except for the ones of Intra-Domain are highlighted with \textbf{bold}.
    }
    \centering
    \resizebox{1\textwidth}{!}{
    \begin{tabular}{cc|cccccccccc|cc}
        \Xhline{1pt}
        \multicolumn{2}{c|}{\multirow{2}*{Methods}} & \multicolumn{2}{c}{Domain 1} & \multicolumn{2}{c}{Domain 2} & \multicolumn{2}{c}{Domain 3} & \multicolumn{2}{c}{Domain 4} & \multicolumn{2}{c|}{Domain 5}  & \multicolumn{2}{c}{Average} \\ 
        \Xcline{3-14}{0.4pt}
        \multicolumn{2}{c|}{} & \multicolumn{2}{c}{DSC$\uparrow$} & \multicolumn{2}{c}{DSC$\uparrow$} & \multicolumn{2}{c}{DSC$\uparrow$} & \multicolumn{2}{c}{DSC$\uparrow$} & \multicolumn{2}{c|}{DSC$\uparrow$} & \multicolumn{2}{c}{DSC$\uparrow$} \\
        \hline
        
        \multicolumn{2}{c|}{Intra-Domain}
        & \multicolumn{2}{c}{$(95.53,82.53)$}
        & \multicolumn{2}{c}{$(94.92,83.94)$}
        & \multicolumn{2}{c}{$(96.08,86.30)$}
        & \multicolumn{2}{c}{$(92.37,85.44)$}
        & \multicolumn{2}{c|}{$(96.07,86.44)$}
        & \multicolumn{2}{c}{$89.96$}
        \\
        \hline
        
        \multicolumn{2}{c|}{DeepAll} 
        & \multicolumn{2}{c}{$(92.87,77.73)$}	
        & \multicolumn{2}{c}{$(91.33,77.41)$}	
        & \multicolumn{2}{c}{$(91.45,79.27)$}	
        & \multicolumn{2}{c}{$(83.51,73.84)$}	
        & \multicolumn{2}{c|}{$(90.82,78.54)$}
        & \multicolumn{2}{c}{$83.68$}
        \\

        \multicolumn{2}{c|}{SAN-SAW (CVPR 2022)~\cite{SAN-SAW}}
        & \multicolumn{2}{c}{$(93.34,76.31)$}  
        & \multicolumn{2}{c}{$(92.88,82.65)$}  
        & \multicolumn{2}{c}{$(90.78,81.18)$}  
        & \multicolumn{2}{c}{$(88.07,77.61)$}  
        & \multicolumn{2}{c|}{$(93.43,83.97)$}
        & \multicolumn{2}{c}{$86.02$}
        \\

        \multicolumn{2}{c|}{DCAC (TMI 2022)~\cite{DCAC}} 
        & \multicolumn{2}{c}{$(94.34,76.72)$}  
        & \multicolumn{2}{c}{$(93.70,79.21)$}  
        & \multicolumn{2}{c}{$(91.05,81.23)$}  
        & \multicolumn{2}{c}{$(88.12,77.87)$}
        & \multicolumn{2}{c|}{$(95.71,85.32)$}  
        & \multicolumn{2}{c}{$86.33$}
        \\

        \multicolumn{2}{c|}{RandConv (ICLR 2021)~\cite{RandConv}} 
        & \multicolumn{2}{c}{$(93.11,76.21)$}  
        & \multicolumn{2}{c}{$(92.50,81.33)$}  
        & \multicolumn{2}{c}{$(89.01,81.33)$}  
        & \multicolumn{2}{c}{$(88.33,76.56)$}  
        & \multicolumn{2}{c|}{$(95.32,85.16)$}  
        & \multicolumn{2}{c}{$85.89$}
        \\

        \multicolumn{2}{c|}{MixStyle (ICLR 2021)~\cite{MixStyle}} 
        & \multicolumn{2}{c}{$(94.40,79.11)$}  
        & \multicolumn{2}{c}{$(92.02,79.19)$}    
        & \multicolumn{2}{c}{$(91.64,80.79)$}    
        & \multicolumn{2}{c}{$(86.41,76.44)$}    
        & \multicolumn{2}{c|}{$(93.09,83.35)$}   
        & \multicolumn{2}{c}{$85.64$}  
        \\

        \multicolumn{2}{c|}{EFDM (CVPR 2022)~\cite{EFDM}}
        & \multicolumn{2}{c}{$(93.84,77.04)$}  
        & \multicolumn{2}{c}{$(91.00,79.53)$}  
        & \multicolumn{2}{c}{$(91.53,81.59)$}  
        & \multicolumn{2}{c}{$(86.31,76.39)$}  
        & \multicolumn{2}{c|}{$(93.69,81.31)$} 
        & \multicolumn{2}{c}{$85.22$}
        \\

        \multicolumn{2}{c|}{DSU (ICLR 2022)~\cite{DSU}}
        & \multicolumn{2}{c}{$(93.71,77.48)$}  
        & \multicolumn{2}{c}{$(91.79,81.65)$}  
        & \multicolumn{2}{c}{$(92.11,79.78)$}  
        & \multicolumn{2}{c}{$(87.96,76.76)$}  
        & \multicolumn{2}{c|}{$(93.58,84.91)$}  
        & \multicolumn{2}{c}{$85.97$}
        \\

        \multicolumn{2}{c|}{MaxStyle (MICCAI 2022)~\cite{MaxStyle}}
        & \multicolumn{2}{c}{$(94.57,77.59)$}  
        & \multicolumn{2}{c}{$(93.67,82.66)$}  
        & \multicolumn{2}{c}{$(\textbf{92.40},79.34)$}  
        & \multicolumn{2}{c}{$(88.81,76.93)$}  
        & \multicolumn{2}{c|}{$(\textbf{96.02},84.39)$}  
        & \multicolumn{2}{c}{$86.64$}
        \\
        \hline
        
        \multicolumn{2}{c|}{TriD (Ours)}
        & \multicolumn{2}{c}{$(\textbf{94.72},\textbf{80.26})$} 	
        & \multicolumn{2}{c}{$(\textbf{93.95},\textbf{82.70})$} 	
        & \multicolumn{2}{c}{$(92.09,\textbf{81.92})$} 	
        & \multicolumn{2}{c}{$(\textbf{90.37},\textbf{78.02})$} 	
        & \multicolumn{2}{c|}{$(95.64,\textbf{86.54})$} 	
        & \multicolumn{2}{c}{$\textbf{87.62}$} 
        \\ 
        \Xhline{1pt}
    \end{tabular}
    }

    \label{tab:OD/OC}
\end{table}

\begin{table}[!tb]
    \caption{Performance of Intra-Domain, DeepAll, our TriD and six DG methods in prostate segmentation. The best results except for the ones of Intra-Domain are highlighted with \textbf{bold}.
    }
    \centering
    \resizebox{1\textwidth}{!}{
    \begin{tabular}{cc|cccccccccccc|cc}
        \Xhline{1pt}
        \multicolumn{2}{c|}{\multirow{2}*{Methods}} & \multicolumn{2}{c}{Domain 1} & \multicolumn{2}{c}{Domain 2} & \multicolumn{2}{c}{Domain 3} & \multicolumn{2}{c}{Domain 4} & \multicolumn{2}{c}{Domain 5} & \multicolumn{2}{c|}{Domain 6}  & \multicolumn{2}{c}{Average} \\ 
        \Xcline{3-16}{0.4pt}
        \multicolumn{2}{c|}{} & DSC$\uparrow$  & ASD $\downarrow$ & DSC$\uparrow$  & ASD $\downarrow$ & DSC$\uparrow$  & ASD $\downarrow$ & DSC$\uparrow$  & ASD $\downarrow$ & DSC$\uparrow$  & ASD $\downarrow$ & DSC$\uparrow$  & ASD $\downarrow$ & DSC$\uparrow$  & ASD $\downarrow$ \\
        \hline
        
        \multicolumn{2}{c|}{Intra-Domain} 	 	
        & $93.24$ & $0.59$
        & $91.85$ & $0.59$
        & $90.52$ & $1.57$
        & $89.69$ & $0.81$
        & $88.19$ & $1.29$
        & $91.09$ & $0.69$
        & $90.76$ & $0.93$
        \\
        \hline

        \multicolumn{2}{c|}{DeepAll} 
        & $90.72$  & $1.04$  
        & $88.53$  & $0.77$  
        & $85.10$  & $3.30$  
        & $88.04$  & $0.91$  
        & $85.84$  & $1.98$  
        & $89.01$  & $0.81$  
        & $87.87$  & $1.47$
        \\

        \multicolumn{2}{c|}{DCAC (TMI 2022)~\cite{DCAC}} 
        & $90.51$  & $0.98$  
        & $88.18$  & $1.34$  
        & $84.35$  & $4.05$ 
        & $88.32$  & $0.83$  
        & $87.01$  & $2.73$  
        & $89.95$  & $0.64$  
        & $88.05$  & $1.76$
        \\

        \multicolumn{2}{c|}{RandConv (ICLR 2021)~\cite{RandConv}} 
        & $90.21$  & $1.01$  
        & $88.59$  & $0.76$  
        & $84.18$  & $3.39$  
        & $88.40$  & $0.73$  
        & $86.80$  & $2.58$  
        & $89.17$  & $0.76$  
        & $87.89$  & $1.54$
        \\

        \multicolumn{2}{c|}{MixStyle (ICLR 2021)~\cite{MixStyle}} 
        & $91.60$  & $0.70$  
        & $90.10$  & $0.69$  
        & $85.62$  & $3.09$  
        & $88.45$  & $0.87$  
        & $87.21$  & $1.45$  
        & $90.02$  & $\textbf{0.61}$  
        & $88.83$  & $1.23$
        \\

        \multicolumn{2}{c|}{EFDM (CVPR 2022)~\cite{EFDM}} 
        & $91.57$  & $0.72$  
        & $90.18$  & $0.70$  
        & $85.34$  & $3.30$  
        & $89.25$  & $0.71$  
        & $86.82$  & $1.71$  
        & $89.52$  & $0.71$  
        & $88.78$  & $1.31$
        \\

        \multicolumn{2}{c|}{DSU (ICLR 2022)~\cite{DSU}} 
        & $90.92$  & $0.79$ 
        & $88.19$  & $0.82$  
        & $84.57$  & $3.86$  
        & $88.68$  & $0.73$  
        & $86.25$  & $1.80$  
        & $89.13$  & $0.70$  
        & $87.96$  & $1.45$
        \\

        \multicolumn{2}{c|}{MaxStyle (MICCAI 2022)~\cite{MaxStyle}} 
        & $90.33$  & $0.79$  
        & $89.17$  & $0.74$  
        & $85.34$  & $2.91$ 
        & $88.72$  & $\textbf{0.67}$  
        & $87.46$  & $1.38$  
        & $88.15$  & $0.72$  
        & $88.19$  & $1.20$
        \\

        \hline
        
        \multicolumn{2}{c|}{TriD (Ours)} 
        & $\textbf{91.63}$  & $\textbf{0.66}$  
        & $\textbf{90.71}$  & $\textbf{0.64}$  
        & $\textbf{86.91}$  & $\textbf{2.77}$  
        & $\textbf{89.42}$  & $0.68$  
        & $\textbf{88.67}$  & $\textbf{1.33}$ 
        & $\textbf{90.11}$  & $0.63$
        & $\textbf{89.57}$  & $\textbf{1.12}$
        
        \\ 
        \Xhline{1pt}
    \end{tabular}
    }

    \label{tab:prostate1}
\end{table}

\noindent{\textbf{Comparing to Other DG methods.}}
We used the same segmentation network and loss function to compare our TriD with seven DG methods, including (1) DCAC: dynamic structure~\cite{DCAC}, (2) SAN-SAW: based on normalization and whitening~\cite{SAN-SAW}, (3) RandConv: input-space domain randomization~\cite{RandConv}, (4-6) MixStyle, EFDM, DSU: feature-space domain randomization~\cite{MixStyle,EFDM,DSU} and (7) MaxStyle: adversarial noise~\cite{MaxStyle}.
Note that since SAN-SAW requires the data with at least two classes, we did not provide its results for prostate segmentation.
Besides, we also compared our method with another two settings, including the `Intra-Domain’ and `DeepAll’. 
Under the `Intra-Domain' setting, training and test data are from the same domain, where three-fold cross-validation is used for prostate segmentation due to the lack of data split. Under the `DeepAll' setting, the model is directly trained on the data aggregated from all source domains and tested on the unseen target domain.
The results are shown in Table~\ref{tab:OD/OC} and Table~\ref{tab:prostate1}. It can be observed that the overall performance of our TriD is not only superior to the `DeepAll' baseline but also better than other DG approaches. 
Furthermore, we found that the performance ranking of MixStyle, EFDM, DSU and our TriD is $TriD > MixStyle \approx EFDM > DSU$, which is consistent with the ranking of search scope in Fig.~\ref{fig-tsne}. 
It reveals that the unreal feature styles are indeed effective, and a larger search space is beneficial to boost the robustness.

\begin{table}[!tb]
    \caption{Performance (OD, OC) of DeepAll, our TriD and its two variants in joint segmentation of OD and OC. The best results are highlighted with \textbf{bold}.
    }
    \centering
    \resizebox{1\textwidth}{!}{
    \begin{tabular}{cc|cccccccccc|cc}
        \Xhline{1pt}
        \multicolumn{2}{c|}{\multirow{2}*{Methods}} & \multicolumn{2}{c}{Domain 1} & \multicolumn{2}{c}{Domain 2} & \multicolumn{2}{c}{Domain 3} & \multicolumn{2}{c}{Domain 4} & \multicolumn{2}{c|}{Domain 5}  & \multicolumn{2}{c}{Average} \\ 
        \Xcline{3-14}{0.4pt}
        \multicolumn{2}{c|}{} & \multicolumn{2}{c}{DSC$\uparrow$} & \multicolumn{2}{c}{DSC$\uparrow$} & \multicolumn{2}{c}{DSC$\uparrow$} & \multicolumn{2}{c}{DSC$\uparrow$} & \multicolumn{2}{c|}{DSC$\uparrow$} & \multicolumn{2}{c}{DSC$\uparrow$} \\
        \hline
        
        \multicolumn{2}{c|}{DeepAll} 
        & \multicolumn{2}{c}{$(92.87,77.73)$}	
        & \multicolumn{2}{c}{$(91.33,77.41)$}	
        & \multicolumn{2}{c}{$(91.45,79.27)$}	
        & \multicolumn{2}{c}{$(83.51,73.84)$}	
        & \multicolumn{2}{c|}{$(90.82,78.54)$}
        & \multicolumn{2}{c}{$83.68$}
        \\
        \hline

        \multicolumn{2}{c|}{DeepAll+SR} 
        & \multicolumn{2}{c}{$(94.30,79.23)$}	
        & \multicolumn{2}{c}{$(91.70,81.15)$}	
        & \multicolumn{2}{c}{$(91.30,81.10)$}	
        & \multicolumn{2}{c}{$(87.87,76.96)$}	
        & \multicolumn{2}{c|}{$(95.08,85.77)$}
        & \multicolumn{2}{c}{$86.45$}
        \\

        \multicolumn{2}{c|}{DeepAll+SR+Mixup} 
        & \multicolumn{2}{c}{$(94.68,78.62)$}	
        & \multicolumn{2}{c}{$(92.08,80.76)$}	
        & \multicolumn{2}{c}{$(91.06,80.81)$}	
        & \multicolumn{2}{c}{$(87.18,75.26)$}	
        & \multicolumn{2}{c|}{$(94.58,84.15)$}
        & \multicolumn{2}{c}{$85.92$}
        \\


        \multicolumn{2}{c|}{TriD (DeepAll+SR+SM)}
        & \multicolumn{2}{c}{$(\textbf{94.72},\textbf{80.26})$} 	
        & \multicolumn{2}{c}{$(\textbf{93.95},\textbf{82.70})$} 	
        & \multicolumn{2}{c}{$(\textbf{92.09},\textbf{81.92})$} 	
        & \multicolumn{2}{c}{$(\textbf{90.37},\textbf{78.02})$} 	
        & \multicolumn{2}{c|}{$(\textbf{95.64},\textbf{86.54})$} 	
        & \multicolumn{2}{c}{$\textbf{87.62}$} 
        \\ 
        \Xhline{1pt}
    \end{tabular}
    }
    \label{tab:ablation}
\end{table}

\noindent{\textbf{Contribution of each component.}}
To evaluate the contribution of statistics randomization (SR) and style mixing (SM), we chose the model trained in joint segmentation of OD and OC as an example and conducted a series of ablation experiments, as shown in Table~\ref{tab:ablation}. 
Note that the `Mixup' denotes the fusion strategy proposed in MixStyle. 
It shows that (1) introducing SR to baseline can lead to huge performance gains; (2) adding the Mixup operation
will degrade the robustness of model due to the limited search space; (3) the best performance is achieved when SR and SM are jointly used (\emph{i.e.}, our TriD).

\begin{figure}[!tb]
	\centering
	\includegraphics[width=\textwidth]{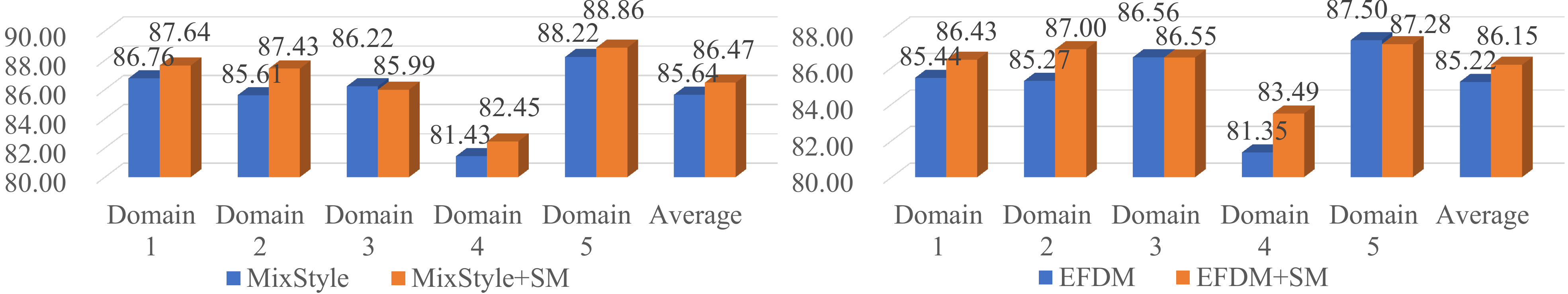}
	\caption{Performance of two DG methods with/without SM in joint segmentation of OD and OC. The value of each domain indicate the mean DSC of OD and OC.}
	\label{fig-SM}
\end{figure}

\noindent{\textbf{Extendibility of SM.}}
Under the same segmentation task, we further evaluate the extendibility of SM by combining it with other DG methods (\emph{i.e.}, MixStyle and EFDM), and the results are shown in Fig.~\ref{fig-SM}. It shows that each approach plus SM achieves better performance in most scenarios and has superior average DSC, proving that our SM strategy can be extended to other DG methods. 

\begin{table}[!tb]
    \caption{Performance (OD, OC) of different locations of TriD in joint segmentation of OD and OC. The best results are highlighted with \textbf{bold}.
    }
    \centering
    \resizebox{1\textwidth}{!}{
    \begin{tabular}{cc|cccccccccc|cc}
        \Xhline{1pt}
        \multicolumn{2}{c|}{\multirow{2}*{Methods}} & \multicolumn{2}{c}{Domain 1} & \multicolumn{2}{c}{Domain 2} & \multicolumn{2}{c}{Domain 3} & \multicolumn{2}{c}{Domain 4} & \multicolumn{2}{c|}{Domain 5}  & \multicolumn{2}{c}{Average} \\ 
        \Xcline{3-12}{0.4pt}
        \multicolumn{2}{c|}{} & \multicolumn{2}{c}{DSC$\uparrow$} & \multicolumn{2}{c}{DSC$\uparrow$} & \multicolumn{2}{c}{DSC$\uparrow$} & \multicolumn{2}{c}{DSC$\uparrow$} & \multicolumn{2}{c|}{DSC$\uparrow$} & \multicolumn{2}{c}{DSC$\uparrow$} \\
        \hline

        \multicolumn{2}{c|}{res1} 
        & \multicolumn{2}{c}{$(94.52,78.12)$}  
        & \multicolumn{2}{c}{$(93.40,\textbf{82.92})$}
        & \multicolumn{2}{c}{$(\textbf{92.14},79.94)$}
        & \multicolumn{2}{c}{$(87.36,77.22)$}
        & \multicolumn{2}{c|}{$(94.84,85.20)$}   
        & \multicolumn{2}{c}{$86.57$}  
        \\
        
        \multicolumn{2}{c|}{res2} 
        & \multicolumn{2}{c}{$(93.65,79.20)$}  
        & \multicolumn{2}{c}{$(93.07,81.16)$}
        & \multicolumn{2}{c}{$(92.12,78.91)$}
        & \multicolumn{2}{c}{$(87.29,77.12)$}
        & \multicolumn{2}{c|}{$(93.47,84.15)$}
        & \multicolumn{2}{c}{$86.01$}  
        \\

        \multicolumn{2}{c|}{res12}
        & \multicolumn{2}{c}{$(\textbf{94.72},\textbf{80.26})$} 	
        & \multicolumn{2}{c}{$(\textbf{93.95},82.70)$} 	
        & \multicolumn{2}{c}{$(92.09,\textbf{81.92})$} 	
        & \multicolumn{2}{c}{$(\textbf{90.37},\textbf{78.02})$} 	
        & \multicolumn{2}{c|}{$(\textbf{95.64},\textbf{86.54})$} 	
        & \multicolumn{2}{c}{$\textbf{87.62}$} 
        \\		 
 	 	 	 	 	 	 	
        \multicolumn{2}{c|}{res123} 
        & \multicolumn{2}{c}{$(92.52,75.11)$}  
        & \multicolumn{2}{c}{$(90.64,80.40)$}
        & \multicolumn{2}{c}{$(91.24,80.06)$}
        & \multicolumn{2}{c}{$(87.82,77.42)$}
        & \multicolumn{2}{c|}{$(95.14,84.18)$}
        & \multicolumn{2}{c}{$85.45$}  
        \\
        
        \multicolumn{2}{c|}{res1234} 
        & \multicolumn{2}{c}{$(90.14,75.27)$}  
        & \multicolumn{2}{c}{$(90.35,79.45)$}
        & \multicolumn{2}{c}{$(91.59,80.76)$}
        & \multicolumn{2}{c}{$(86.77,77.04)$}
        & \multicolumn{2}{c|}{$(94.21,84.17)$}
        & \multicolumn{2}{c}{$84.97$}  
        \\
        \Xhline{1pt}
    \end{tabular}
    }

    \label{tab:Location}
\end{table}

\noindent{\textbf{Location of TriD.}}
To discuss where to apply our TriD, we repeated the experiments in joint segmentation of OD and OC and listed the results in Table~\ref{tab:Location}. These four residual blocks of ResNet-34 are denoted as `res1-4', and we trained different variants via applying TriD to different blocks. 
The results reveal that (1) the best performance is achieved by applying TriD to `res12' that extract the low-level features with the most style-texture information; (2) the performance degrades when applying TriD to the third and last blocks that tend to capture semantic content rather than style texture~\cite{IBN}.


\begin{table}[!tb]
    \caption{Performance (OD, OC) of using normal distribution and uniform distribution in joint segmentation of OD and OC. The best results are highlighted with \textbf{bold}.
    }
    \centering
    \resizebox{1\textwidth}{!}{
    \begin{tabular}{cc|cccccccccc|cc}
        \Xhline{1pt}
        \multicolumn{2}{c|}{\multirow{2}*{Methods}} & \multicolumn{2}{c}{Domain 1} & \multicolumn{2}{c}{Domain 2} & \multicolumn{2}{c}{Domain 3} & \multicolumn{2}{c}{Domain 4} & \multicolumn{2}{c|}{Domain 5}  & \multicolumn{2}{c}{Average} \\ 
        \Xcline{3-12}{0.4pt}
        \multicolumn{2}{c|}{} & \multicolumn{2}{c}{DSC$\uparrow$} & \multicolumn{2}{c}{DSC$\uparrow$} & \multicolumn{2}{c}{DSC$\uparrow$} & \multicolumn{2}{c}{DSC$\uparrow$} & \multicolumn{2}{c|}{DSC$\uparrow$} & \multicolumn{2}{c}{DSC$\uparrow$} \\
        \hline

        \multicolumn{2}{c|}{Normal Distribution} 
        & \multicolumn{2}{c}{$(92.93,78.22)$}  
        & \multicolumn{2}{c}{$(91.37,81.06)$}    
        & \multicolumn{2}{c}{$(91.64,80.60)$}    
        & \multicolumn{2}{c}{$(84.66,72.20)$}    
        & \multicolumn{2}{c|}{$(93.55,83.39)$}   
        & \multicolumn{2}{c}{$84.96$}  

        \\

        \multicolumn{2}{c|}{Uniform Distribution}
        & \multicolumn{2}{c}{$(\textbf{94.72},\textbf{80.26})$} 	
        & \multicolumn{2}{c}{$(\textbf{93.95},\textbf{82.70})$} 	
        & \multicolumn{2}{c}{$(\textbf{92.09},\textbf{81.92})$} 	
        & \multicolumn{2}{c}{$(\textbf{90.37},\textbf{78.02})$} 	
        & \multicolumn{2}{c|}{$(\textbf{95.64},\textbf{86.54})$} 	
        & \multicolumn{2}{c}{$\textbf{87.62}$} 
        \\ 
        \Xhline{1pt}
    \end{tabular}
    }

    \label{tab:distribution}
\end{table}

\noindent{\textbf{Uniform distribution vs. normal distribution.}}
It is particularly critical to choose the distribution from which to randomly sample the augmented statistics. To verify the advantages of uniform distribution, we repeated the experiments in joint segmentation of OD and OC by replacing the uniform distribution with a normal distribution $N(0.5, 1)$ and compared the effectiveness of them in Table~\ref{tab:distribution}. It shows that the normal distribution indeed results in performance drop due to the limited search space.


\section{Conclusion}
We proposed the TriD, a domain-randomization based multi-source domain generalization method, for medical image segmentation.
To solve the limitations existing in preview methods, TriD perturbs the intermediate features with two steps: (1) SR: randomly sampling the augmented statistics from a uniform distribution to expand the search space of feature styles; (2) SM: mixing the feature styles for explicit domain-invariant representation learning.
Through extensive experiments on two medical segmentation tasks with different modalities, 
the proposed TriD is demonstrated to achieve superior performance over the baselines and other state-of-the-art DG methods.

\subsubsection{Acknowledgment}
This work was supported in part by the National Natural Science Foundation of China under Grants 62171377, in part by the National Key R\&D Program of China under Grant 2022YFC2009903 / 2022YFC2009900, in part by the Key Research and Development Program of Shaanxi Province, China, under Grant 2022GY-084, and in part by the China Postdoctoral Science Foundation BX2021333 / 2021M703340.


%
%
%

\bibliographystyle{splncs04}
\bibliography{references}

\newpage
\section{Appendix}

\begin{algorithm}[H]
\caption{PyTorch-like pseudo-code for TriD.}
\begin{minted}{python}
# x: input intermediate features (B, C, H, W)
# p: probability to apply TriD (default: 0.5)
# alpha: hyper-parameter for the Beta distribution (default: 0.1)
# eps: a small value added before square root (default: 1e-6)
if random probability > p:
    return x
B, C, H, W = x.shape
mu, var = x.mean(dim=[2,3], keepdim=True), x.var(dim=[2,3], keepdim=True)
sig = (var + eps).sqrt()
x_normed = (x - mu.detach()) / sig.detach() # normalize the features
mu_r = torch.empty((B, C, 1, 1)).uniform_(0.0, 1.0) # sample the mu
sig_r = torch.empty((B, C, 1, 1)).uniform_(0.0, 1.0) # sample the std
lmda = torch.distributions.Beta(alpha, alpha).sample((B, C, 1, 1))
bern = torch.bernoulli(lmda) # binarization
mu_mix = mu_r * bern + mu * (1. - bern) # generate the mixed mu
sig_mix = sig_r * bern + sig * (1. - bern) # generate the mixed std
return x_normed * sig_mix + mu_mix # perturb the features
\end{minted}
\end{algorithm}

\begin{table}[!htb]
    \caption{Performance of our TriD and five state-of-the-art methods in prostate segmentation under the leave-one-domain-out setting. All the methods used nnUNet as the backbone. The results except for the ones of our TriD are adopted from the paper of DCAC. The best results are highlighted with \textbf{bold}.
    }
    \centering
    \resizebox{1\textwidth}{!}{
    \begin{tabular}{cc|cccccccccccc|cc}
        \Xhline{1pt}
        \multicolumn{2}{c|}{\multirow{2}*{Methods}} & \multicolumn{2}{c}{Domain 1} & \multicolumn{2}{c}{Domain 2} & \multicolumn{2}{c}{Domain 3} & \multicolumn{2}{c}{Domain 4} & \multicolumn{2}{c}{Domain 5} & \multicolumn{2}{c|}{Domain 6}  & \multicolumn{2}{c}{Average} \\ 
        \Xcline{3-16}{0.4pt}
        \multicolumn{2}{c|}{} & DSC$\uparrow$  & ASD $\downarrow$ & DSC$\uparrow$  & ASD $\downarrow$ & DSC$\uparrow$  & ASD $\downarrow$ & DSC$\uparrow$  & ASD $\downarrow$ & DSC$\uparrow$  & ASD $\downarrow$ & DSC$\uparrow$  & ASD $\downarrow$ & DSC$\uparrow$  & ASD $\downarrow$ \\
        \hline
    \multicolumn{2}{c|}{BigAug} & $90.68$ & $1.80$ & $89.52$ & $1.00$ & $84.86$ & $1.86$ & $89.04$ & $1.59$ & $73.24$ & $5.94$ & $89.10$ & $1.16$ & $86.07$ & $2.23$  \\
    \multicolumn{2}{c|}{SAML} & $91.00$ & $1.26$ & $89.26$ & $1.12$ & $85.76$ & $1.87$ & $\textbf{89.60}$ & $1.21$ & $81.60$ & $3.29$ & $89.91$ & $0.96$ & $87.86$ & $1.62$  \\
    \multicolumn{2}{c|}{FedDG} & $91.41$ & $1.29$ & $89.95$ & $0.97$ & $85.10$ & $2.63$ & $89.13$ & $1.51$ & $76.69$ & $4.52$ & $90.63$ & $1.03$ & $87.15$ & $1.99$  \\
    \multicolumn{2}{c|}{DoFE} & $89.79$ & $1.33$ & $87.42$ & $1.57$ & $84.90$ & $2.13$ & $88.56$ & $1.52$ & $\textbf{86.47}$ & $1.93$ & $87.72$ & $1.33$ & $87.48$ & $1.64$  \\
    \multicolumn{2}{c|}{DCAC} & $91.76$ & $0.98$ & $90.51$ & $0.89$ & $86.30$ & $\textbf{1.77}$ & $89.13$ & $1.53$ & $83.39$ & $2.46$ & $90.56$ & $0.85$ & $88.61$ & $1.41$  \\
    \multicolumn{2}{c|}{TriD (Ours)} & $\textbf{92.30}$ & $\textbf{0.66}$ & $\textbf{90.56}$ & $\textbf{0.67}$ & $\textbf{86.72}$ & $2.71$ & $89.16$ & $\textbf{0.74}$ & $85.60$ & $\textbf{1.79}$ & $\textbf{91.57}$ & $\textbf{0.56}$ & $\textbf{89.32}$ & $\textbf{1.19}$  \\

        \Xhline{1pt}
    \end{tabular}
}

    \label{tab:prostate2}
\end{table}

\begin{figure}[!htb]
	\centering
	\includegraphics[width=\textwidth]{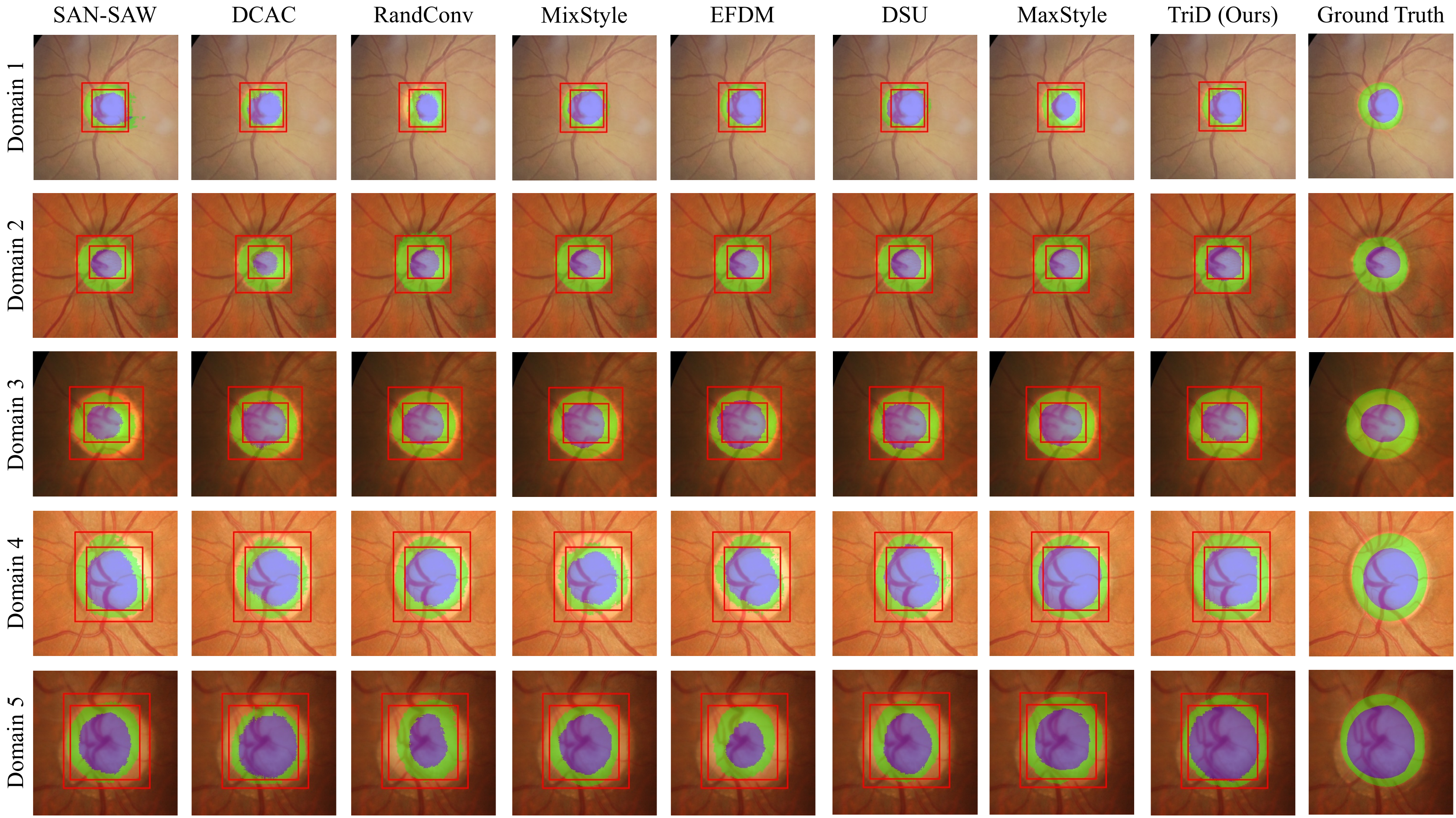}
	\caption{Visualization results of seven state-of-the-art methods and our TriD in joint segmentation of OD and OC under the leave-one-domain-out setting. The ground-truth bounding boxes of OD and OC were overlaid on each segmentation result to highlight possible over- or under-segmentation. Best viewed in color.}
	\label{fig-seg}
\end{figure}

\begin{table}[!htb]
    \caption{Since our TriD can also be applied to single-source domain generalization, we conducted the experiments to compare our TriD with other methods under the single-source setting, \emph{i.e.}, training on one source domain and evaluating on other domains. Note that the results of DCAC are not provided since it relies on multi-source data.
    Performance (OD, OC) is shown as follow. The best results are highlighted with \textbf{bold}.
    }
    \centering
    \resizebox{1\textwidth}{!}{
    \begin{tabular}{cc|cccccccccc|cc}
        \Xhline{1pt}
        \multicolumn{2}{c|}{\multirow{2}*{Methods}} & \multicolumn{2}{c}{Domain 1} & \multicolumn{2}{c}{Domain 2} & \multicolumn{2}{c}{Domain 3} & \multicolumn{2}{c}{Domain 4} & \multicolumn{2}{c|}{Domain 5}  & \multicolumn{2}{c}{Average} \\ 
        \Xcline{3-14}{0.4pt}
        \multicolumn{2}{c|}{} & \multicolumn{2}{c}{DSC$\uparrow$} & \multicolumn{2}{c}{DSC$\uparrow$} & \multicolumn{2}{c}{DSC$\uparrow$} & \multicolumn{2}{c}{DSC$\uparrow$} & \multicolumn{2}{c|}{DSC$\uparrow$} & \multicolumn{2}{c}{DSC$\uparrow$} \\
        \hline

    \multicolumn{2}{c|}{DeepAll} & 
    \multicolumn{2}{c}{$(74.54, 59.21)$}  & 
    \multicolumn{2}{c}{$(82.30, 71.96)$}  & 
    \multicolumn{2}{c}{$(78.06, 59.12)$}  & 
    \multicolumn{2}{c}{$(79.79, 59.23)$}  & 
    \multicolumn{2}{c|}{$(85.25, 58.88)$}  & 
    \multicolumn{2}{c}{$70.83$}  
    \\    
    \multicolumn{2}{c|}{SAN-SAW (CVPR 2022)} & 
    \multicolumn{2}{c}{$(76.42, 59.01)$}  & 
    \multicolumn{2}{c}{$(83.79, 73.23)$}  & 
    \multicolumn{2}{c}{$(84.17, 65.51)$}  & 
    \multicolumn{2}{c}{$(81.83, 62.36)$}  & 
    \multicolumn{2}{c|}{$(87.00, 64.42)$}  & 
    \multicolumn{2}{c}{$73.77$}  
    \\
    \multicolumn{2}{c|}{RandConv (ICLR 2022)} & 
    \multicolumn{2}{c}{$(79.63, 64.14)$}  & 
    \multicolumn{2}{c}{$(85.00, 72.40)$}  & 
    \multicolumn{2}{c}{$(87.77, 69.57)$}  & 
    \multicolumn{2}{c}{$(83.08, 64.38)$}  & 
    \multicolumn{2}{c|}{$(86.31, 60.37)$}  & 
    \multicolumn{2}{c}{$75.27$}  
    \\
    \multicolumn{2}{c|}{MixStyle (ICLR 2021)} & 
    \multicolumn{2}{c}{$(75.67, 60.84)$}  & 
    \multicolumn{2}{c}{$(86.35, 73.77)$}  & 
    \multicolumn{2}{c}{$(85.86, 66.60)$}  & 
    \multicolumn{2}{c}{$(\textbf{84.86}, 66.44)$}  & 
    \multicolumn{2}{c|}{$(86.54, 65.99)$}  & 
    \multicolumn{2}{c}{$75.29$}  
    \\
    \multicolumn{2}{c|}{EFDM (CVPR 2022)} & 
    \multicolumn{2}{c}{$(78.79, 57.73)$}  & 
    \multicolumn{2}{c}{$(84.83, 72.20)$}  & 
    \multicolumn{2}{c}{$(85.25, 65.94)$}  & 
    \multicolumn{2}{c}{$(82.13, 61.62)$}  & 
    \multicolumn{2}{c|}{$(85.45, 63.02)$}  & 
    \multicolumn{2}{c}{$73.70$}  
    \\
    \multicolumn{2}{c|}{DSU (ICLR 2022)} & 
    \multicolumn{2}{c}{$(76.88, 61.26)$}  & 
    \multicolumn{2}{c}{$(84.17, 74.10)$}  & 
    \multicolumn{2}{c}{$(89.12, 70.16)$}  & 
    \multicolumn{2}{c}{$(83.53, 63.19)$}  & 
    \multicolumn{2}{c|}{$(87.09, 59.65)$}  & 
    \multicolumn{2}{c}{$74.91$}  
    \\
    \multicolumn{2}{c|}{MaxStyle (MICCAI 2022)} & 
    \multicolumn{2}{c}{$(77.40, 65.44)$}  & 
    \multicolumn{2}{c}{$(86.95, 74.52)$}  & 
    \multicolumn{2}{c}{$(87.95, 67.62)$}  & 
    \multicolumn{2}{c}{$(84.69, 66.05)$}  & 
    \multicolumn{2}{c|}{$(\textbf{87.95}, 64.84)$}  & 
    \multicolumn{2}{c}{$76.34$}  
    \\
    \multicolumn{2}{c|}{TriD (Ours)} & 
    \multicolumn{2}{c}{$(\textbf{81.86}, \textbf{66.67})$}  & 
    \multicolumn{2}{c}{$(\textbf{88.19}, \textbf{75.43})$}  & 
    \multicolumn{2}{c}{$(\textbf{89.62}, \textbf{70.85})$}  & 
    \multicolumn{2}{c}{$(84.81, \textbf{67.53})$}  & 
    \multicolumn{2}{c|}{$(87.88, \textbf{66.96})$}  & 
    \multicolumn{2}{c}{$\textbf{77.98}$}  
    \\
        \Xhline{1pt}
    \end{tabular}
    }

    \label{tab:OD/OC2}
\end{table}

\end{document}